\documentclass[10pt,twocolumn,letterpaper]{article}

\usepackage{iccv}
\usepackage{times}
\usepackage{epsfig}
\usepackage{graphicx}
\usepackage{amsmath}
\usepackage{amssymb}

\usepackage{textcomp}
\usepackage[inline]{enumitem}
\usepackage{gensymb}
\usepackage{pdfpages}
\usepackage{tabularx}
\usepackage{caption}
\usepackage{subcaption}
\usepackage{float}

\usepackage[breaklinks=true,bookmarks=false]{hyperref}

\usepackage{cleveref}

\iccvfinalcopy 

\def\iccvPaperID{5486} 

\begin{document}

\title{Fast 3D Pose Refinement with RGB Images}

\author{Abhinav Jain and Frank Dellaert\\
Georgia Institute of Technology\\
Atlanta, GA 30332, USA\\
{\tt\small \{jain, fd27\}@gatech.edu}
}

\maketitle

\begin{abstract}
   Pose estimation is a vital step in many robotics and perception tasks such as robotic manipulation, autonomous vehicle navigation, etc. Current state-of-the-art pose estimation methods rely on deep neural networks with complicated structures and long inference times. While highly robust, they require computing power often unavailable on mobile robots. We propose a CNN-based pose refinement system which takes a coarsely estimated 3D pose from a computationally cheaper algorithm along with a bounding box image of the object, and returns a highly refined pose. Our experiments on the YCB-Video dataset show that our system can refine 3D poses to an extremely high precision with minimal training data.
\end{abstract}
\vspace{-10pt}
\section{Introduction}
In this work, we focus on the challenging task of 3D rotation estimation -- rotational orientation of an object with respect to a given reference frame, usually the camera --, an important topic in computer vision.
The 6-DOF pose includes the 3D translation of the object in the scene, and the 3D rotation of the object in the scene. A given 3D scene can be represented by 6-DOF poses of all objects in the scene. 
Pose estimation is a vital step in many robotics pipelines such as robotic manipulation and autonomous driving, but current methods have high computational requirements. While some recent pose estimation methods such as \cite{tekin2018real} and \cite{kehl2017ssd} have been able to achieve real time performance, they require high-end GPUs for fast inference. Their performance on CPUs is significantly limited, even though mobile robots or other embedded devices generally have limited computing hardware due to power and size constraints.  
Another issue with state-of-the-art deep neural networks is that they require large training datasets to achieve good performance, often in the order of several hundreds of thousands of samples. However, annotating 6D object pose datasets is an extremely challenging task. For example, to create the popular LINEMOD object pose dataset \cite{hinterstoisser2012model}, each frame was annotated by using a robotic arm to control the camera, thereby providing the ground truth rotation for each frame. Despite this the dataset is still limited to just 1,000 images for each of the 15 object classes, which is significantly less than what would normally be required to train a deep neural network. As an alternative solution, creating synthetic images using 3D models of the objects is an option. However, performance of a neural network trained on computer generated images may not generalize to the real world.
The key idea in this paper is to propose a two-stage approach: instead of estimating the pose from scratch, we assume that, along with the location of the object, a coarse estimate of the 3D pose of the object is provided to us. Given an RGB image of an object along with its location in the image represented as a bounding box around the object, we attempt to provide an estimate of the 3D rotation of the object with respect to the camera. 
Our key contribution is a shallow neural network framework that is able to use a bounding box RGB image of the object along with a course 3D pose estimate to produce a highly precise 3D pose. Our system relies only on the RGB bounding box image, and does not require a 3D CAD model of the object for training unlike other works that have attempted this solution.
Our system can be trained on smaller datasets, and can provide quick inference on computationally limited hardware. 
One advantage of using a shallow neural network for refinement is the fewer number of training samples required to achieve optimal performance. 
Our network provides optimal performance with just a few thousand labeled samples per object, and allows for easy dataset augmentation without the use of synthetic images.
By using a shallow neural network for regression of the refined 3D pose, we significantly reduce the computing cost of the 3D pose estimation pipeline. Thus, mobile robots can take advantage of a shallow network based pose refinement algorithm in addition to complement a fast but coarse pose estimation system.
Refining an estimated 3D pose has other significant advantages over an end-to-end pose estimation framework. It provides flexibility to refine a pose to a certain degree while taking a tradeoff between precision and inference time into account, and can be used in iterative optimization methods.

\section{Related Works}

3D pose estimation and object detection are well explored topics. Literature in this domain includes research on 6D pose estimation. In this section, we will focus on the more recent work on this problem. We will also limit our review to work that involves pose estimation using convolutional neural networks on single RGB images rather than methods that use more complex sensing techniques. Further, we will review both general 3D object pose methods as well as methods that specifically involve pose refinement as an optional final step.

2D keypoint detection and matching are commonly used for 3D pose estimation from RGB images. Keypoints in the 2D image are detected using methods such as SIFT \cite{lowe1999object} and matched to keypoints in 3D models of the objects. Then, the projection of the 3D keypoints are aligned to the 2D keypoints to estimate the pose of the object. This method, used by \cite{wu2016single} and \cite{pavlakos20176}, is quite computationally demanding. 

Newer methods attempt to estimate the 3D pose directly from an RGB image without the use of a 3D model. \cite{mahendran20173d} regresses on the 3D pose of objects in the 3D rotation space. Their pipeline takes in bounding box images of un-occluded and un-truncated objects, and outputs a pose as an axis-angle rotation or a unit quaternion (see \cref{pose_representation}). Unlike earlier 3D pose estimation methods such as \cite{su2015render} and \cite{tulsiani2015viewpoints} that approach pose estimation as a classification problem, wherein they classify rotations into bins, \cite{mahendran20173d} attempt to directly regress on the 3D rotation of the object. 

Pose refinement as a secondary stage to pose estimation in order to improve results was first used in \cite{oberweger2015training} for hand pose estimation. They used a synthetic image of the hand created using the estimated pose from the first stage to update the pose estimate in an iterative algorithm.

\cite{rad2017bb8, kehl2017ssd, manhardt2018deep} subsequently used the refinement approach in 3D object pose estimation. \cite{rad2017bb8} present an end-to-end 3D object pose estimation system using a deep convolutional neural network on RGB images. Their optional pose refinement stage involves generating a binary mask or a color render of the object using a 3D model of the object at the estimated pose, and using a CNN to minimize the error between the projection of the 3D bounding box of the ground truth object pose and the newly rendered object pose. \cite{kehl2017ssd} also render each object into the scene using the estimated pose, and then try to minimize the projection error of contours of the rendered surfaces against the ground truth. \cite{manhardt2018deep} propose a similar approach. After a course initial 3D pose hypothesis, they iteratively minimize the projection from the rendered models till they converge to a refined pose. All these methods use a 3D CAD model of the object for pose refinement. Our method is the first to refine a pose using just the RGB image.

\section{Representing 3D Pose}
\label{pose_representation}

The 3D orientation of an object in the world is commonly represented by Euler angles. However, Euler angles are not an ideal representation of rotation since they suffer from singularities. We can instead use an axis-angle representation, i.e., the orientation is represented as a 3D rotation by an angle $\theta$ about a given axis $\hat{n}$ passing through the origin of the object, as calculated in the object localization task. This representation of a rotation is called the axis-angle representation. A 3D rotation can also be represented by a rotation matrix, which lies in the set $SO(3)$ of the special orthogonal matrices of dimension 3, i.e., matrices that are orthogonal, and with determinant $1$:
\begin{equation}
   SO(3) = \left\{R:R\in\mathbb{R}^{3\times3},R^{T}R=I_{3},\det\left(R\right)=1\right\}   
\end{equation}
Another popular representation of 3D rotations is unit quaternions. There is a one-one translations between rotation matrices and unit quaternions. For an axis-angle rotation represented by unit axis $\hat{n}=(\hat{n}_{x},\hat{n}_{y},\hat{n}_{z})$ and angle $\theta$, the corresponding unit quaternion is:
\begin{equation}
   \label{eqn:axisangle_to_quat}
   q\left(\hat{n},\theta\right) = \left[\cos\frac{\theta}{2},n_{x}\sin\frac{\theta}{2},n_{y}\sin\frac{\theta}{2},n_{z}\sin\frac{\theta}{2}\right].
\end{equation}
Similarly, for a given quaternion $q=\left(q_{w},q_{x},q_{y},q_{z}\right)$, the corresponding axis-angle rotation would be:
\begin{equation}
   \theta=2\arccos\left(q_{w}\right),\;n=\frac{1}{{\sqrt{1-q_{w}^{2}}}}  \left[q_x, q_y, q_z\right]
\end{equation}

A quaternion can be split into a scalar part $r$ and a vector part $\vec{v}$. That is:
\begin{equation}
   q = \left(r, \vec{v}\right)\text{, where } r = q_w \text{ and } \vec{v} = \left(q_x, q_y, q_z\right)
\end{equation}
With this, we can perform multiplication for quaternions using the standard vector cross and dot products. Multiplication of two quaternions is the same as performing the rotation operations represented by each quaternion sequentially.
\begin{multline}
      q_1 * q_2 = \left(r_1, \vec{v}_1\right)\left(r_2, \vec{v}_2\right)\\
      = \left(r_1 r_2 - \vec{v}_1\cdot\vec{v}_2,r_1\vec{v}_2+r_2\vec{v}_1+\vec{v}_1\times\vec{v}_2\right)
\end{multline}
We also define the quaternion $q^{-1}$ as the quaternion obtained by using $-\theta$ in \cref{eqn:axisangle_to_quat}.
\begin{equation}
   q\left(\hat{n},\theta\right) = \left(r,\vec{v}\right),\; q^{-1} = q\left(\hat{n}, -\theta\right) = \left(r,-\vec{v}\right)
\end{equation}
This represents the reverse of the original rotation. Rotation by $-\theta$ around axis $\hat{n}$ is the same as rotating by angle $\theta$ around axis $-\hat{n}$. From this, we can define the geodesic angle between two quaternions:
\begin{equation}
   \theta_{geodesic} = 2\arccos\left(\left|q_1 * q_2^{-1}\right|\right)
   \label{eqn:geodesic}
\end{equation}
where $\left|q\right| = \left|\left(r,\vec{v}\right)\right| = r$ is the scalar component of a quaternion.

\section{Dataset}

\begin{figure}
   \begin{center}
      \includegraphics[width=\linewidth]{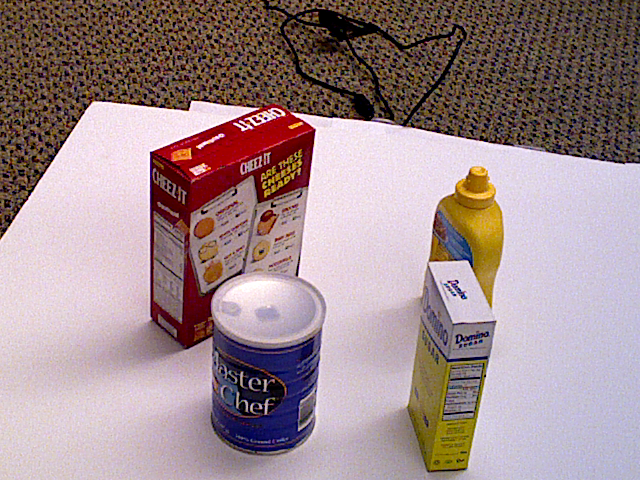}
   \end{center}
   \caption{Sample Frame from the YCB-Video Dataset}
   \label{fig:data_sample}
\end{figure}

We used the YCB-Video dataset \cite{xiang2018posecnn} for all our testing and training. The dataset consists of individual frames from video clips of 21 common objects. Each frame is annotated with ground truth object pose as a 3D rotation matrix as well as bounding box coordinates for each object in the frame. Other annotations such as extrinsic camera parameters and object 3D translation were provided as well but were not required for our framework. \Cref{fig:data_sample} is a sample image from the dataset.

The rounding error while reading the rotation matrix for the ground truth pose meant resulted in non-orthogonal rotational matrices. This lead to singularities when converting the rotation matrices to unit quaternions. Thus, we first had to re-orthogonalize the rotation matrix. We used a method recommended in \cite{premerlani2009direction}. We first took two of the three orthogonal vectors from the rotation matrix and evaluated the error in their dot product:
\begin{equation}
   \mathbf{X}=\begin{bmatrix}r_{xx}\\
   r_{xy}\\
   r_{xz}
   \end{bmatrix},\text{ }\mathbf{Y}=\begin{bmatrix}r_{yx}\\
   r_{yy}\\
   r_{xz}
   \end{bmatrix}
\end{equation}
\begin{equation}
   error=\mathbf{X}\cdot\mathbf{Y}   
\end{equation}
We used half of the error to re-orthogonalize the two vectors, and create the third vector as the cross product of the two re-orthogonalized vectors:
\begin{equation}
   \mathbf{X}_{orthogonal}=\mathbf{X}-\frac{error}{2}\mathbf{Y}
\end{equation}
\begin{equation}
   \mathbf{Y}_{orthogonal}=\mathbf{Y}-\frac{error}{2}\mathbf{X}
\end{equation}
\begin{equation}
   \mathbf{Z}_{orthogonal}=\mathbf{X}_{orthogonal}\times\mathbf{Y}_{orthogonal}   
\end{equation}
Finally, we normalized each of the three orthogonal vectors:
\begin{equation}
   \mathbf{X}_{normalized} = \frac{1}{2}\left(3 - \mathbf{X}_{orthogonal}\cdot\mathbf{X}_{orthogonal}\right)\mathbf{X}_{orthogonal}
\end{equation}

\begin{figure*}
   \centering
   \begin{subfigure}[t]{0.1\linewidth}
      \centering
      \includegraphics[height = 1in]{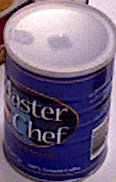}     
      \caption{Tin Can}    
   \end{subfigure}%
   ~
   \begin{subfigure}[t]{0.1\linewidth}
      \centering
      \includegraphics[height = 1in]{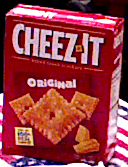}   
      \caption{Crackers}      
   \end{subfigure}%
   ~
   \begin{subfigure}[t]{0.1\linewidth}
      \centering
      \includegraphics[height = 1in]{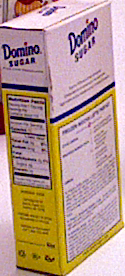}      
      \caption{Sugar Box}   
   \end{subfigure}%
   ~
   \begin{subfigure}[t]{0.1\linewidth}
      \centering
      \includegraphics[height = 1in]{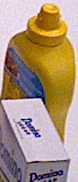}
      \caption{Mustard}
   \end{subfigure}%
   ~
   \begin{subfigure}[t]{0.1\linewidth}
      \centering
      \includegraphics[height = 1in]{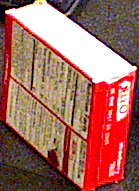}
      \caption{Gelatin}
   \end{subfigure}%
   ~
   \begin{subfigure}[t]{0.1\linewidth}
      \centering
      \includegraphics[width=\linewidth]{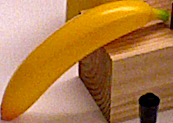}
      \caption{Banana}
   \end{subfigure}%
   ~
   \begin{subfigure}[t]{0.1\linewidth}
      \centering
      \includegraphics[height = 1in]{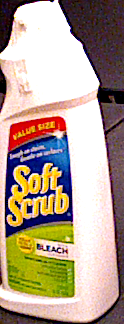}
      \caption{Bleach}
   \end{subfigure}%
   ~
   \begin{subfigure}[t]{0.1\linewidth}
      \centering
      \includegraphics[height = 1in]{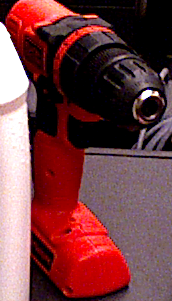}
      \caption{Drill}
   \end{subfigure}%
   ~
   \begin{subfigure}[t]{0.1\linewidth}
      \centering
      \includegraphics[width=\linewidth]{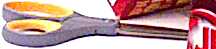}
      \caption{Scissors}
   \end{subfigure}%
   \caption{Objects from the YCB-Video Dataset}
   \label{fig:objects}
\end{figure*}

We used 9 of the 21 objects for all our experiments. \Cref{fig:objects} shows the 9 objects used. We had over 12,000 bounding box images and corresponding ground truth poses for each of the 9 objects. We used 5,000 samples for training, 3,000 for validation, and 3,000 for testing. Further, since the frames were recorded as video clips, pose change between subsequent frames was incremental. Thus, each object database was randomized in order.

\section{Network Architecture}

\begin{figure*}
   \begin{center}
      \includegraphics[width=\linewidth]{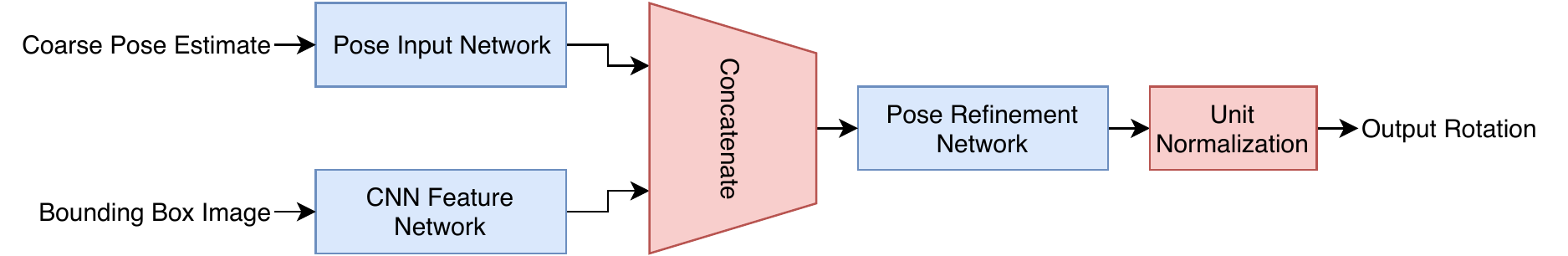}
   \end{center}
   \caption{Overview of our pose refinement pipeline}
   \label{fig:architecture}
\end{figure*}

Our network is composed of three stages -- the CNN feature extraction stage, the fully connected pose input stage, and the fully connected pose refinement stage. The feature extraction stage takes in the bounding box image of the object as an input. The input image is resized for the network. The pose input stage takes a course estimate of the pose as a unit quaternion as input. The outputs of both these stages are concatenated and input into the pose refinement stage. The final 4 dimensional output is unit normalized. \Cref{fig:architecture} shows the overall architecture of the pose refinement pipeline.

Instead of directly outputting a refined 3D pose, we output a unit quaternion representing the rotation from the input pose estimate to the ground truth. The reason for this choice is to prevent the network from over-fitting to the image data. The network should factor both the bounding box image as well as the coarse pose estimate in the final output. We explored this further in \cref{network_invarience}.


\subsection{CNN Feature Extraction Stage}

\begin{figure}[H]
   \begin{center}
      \includegraphics[width=0.5\linewidth]{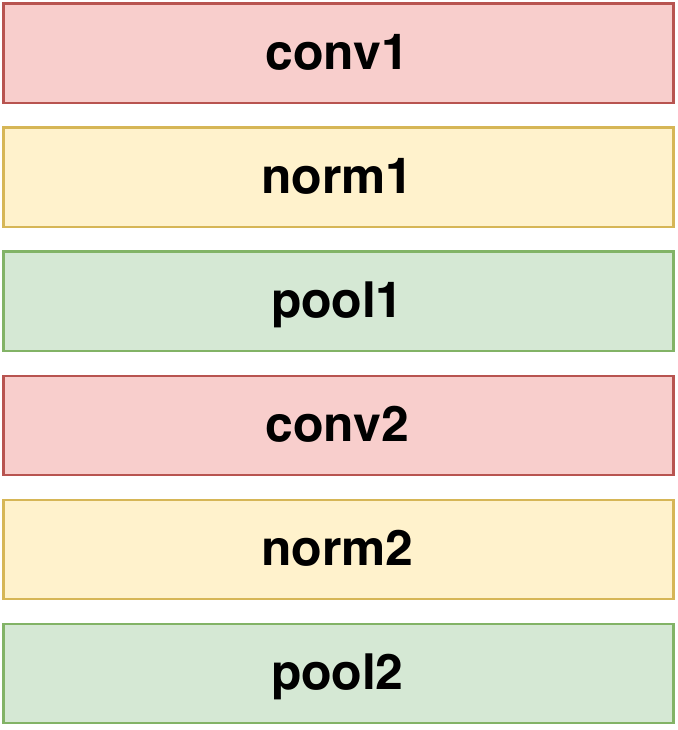}
   \end{center}
   \caption{Architecture of the CNN feature extraction stage}
   \label{fig:cnn_stage}
\end{figure}

The CNN feature extraction stage is a simple shallow CNN network composed of two convolutional layers, two batch normalization layers, and two max pooling layers. \Cref{fig:cnn_stage} shows the architecture of the stage. While deeper CNN networks perform really well in classification tasks, 3D pose estimation relies mostly on finer image features extracted by the first few layers of CNN networks. This is further true for pose refinement. Thus, just these few layers are adequate for our application. We verified this by comparing our CNN network against the VGG-M network in our framework in \cref{exp_CNN}.

\subsection{Fully Connected Pose Input Stage}

\begin{figure}
   \begin{center}
      \includegraphics[width=0.5\linewidth]{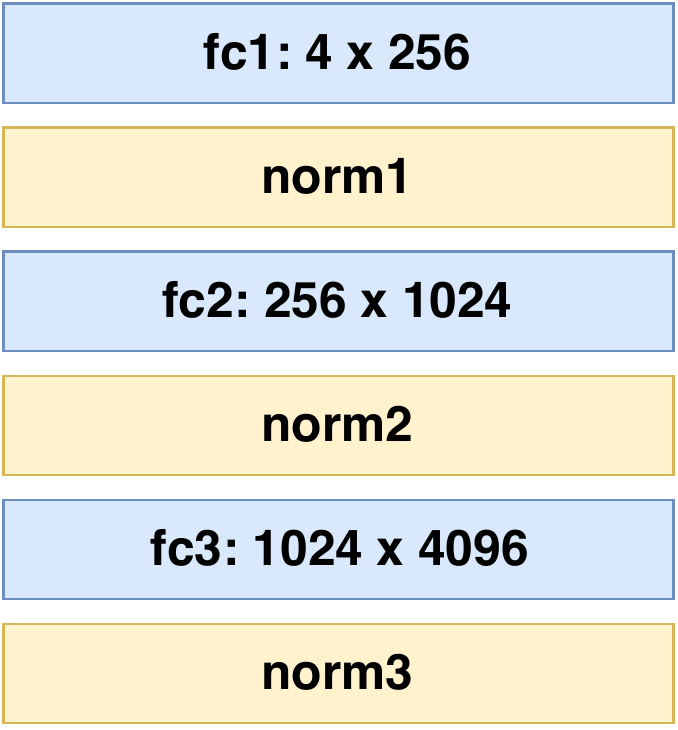}
   \end{center}
   \caption{Architecture of the pose input stage}
   \label{fig:pose_input_stage}
\end{figure}

The fully connected pose input stage consists of 3 fully connected layers that expand the dimension of the input pose estimate from 4 to 4,096. This matches the output dimension of the CNN features extraction stage. \Cref{fig:pose_input_stage} shows the architecture of this stage.

\subsection{Pose Refinement Stage}

\begin{figure}
   \begin{center}
      \includegraphics[width=0.5\linewidth]{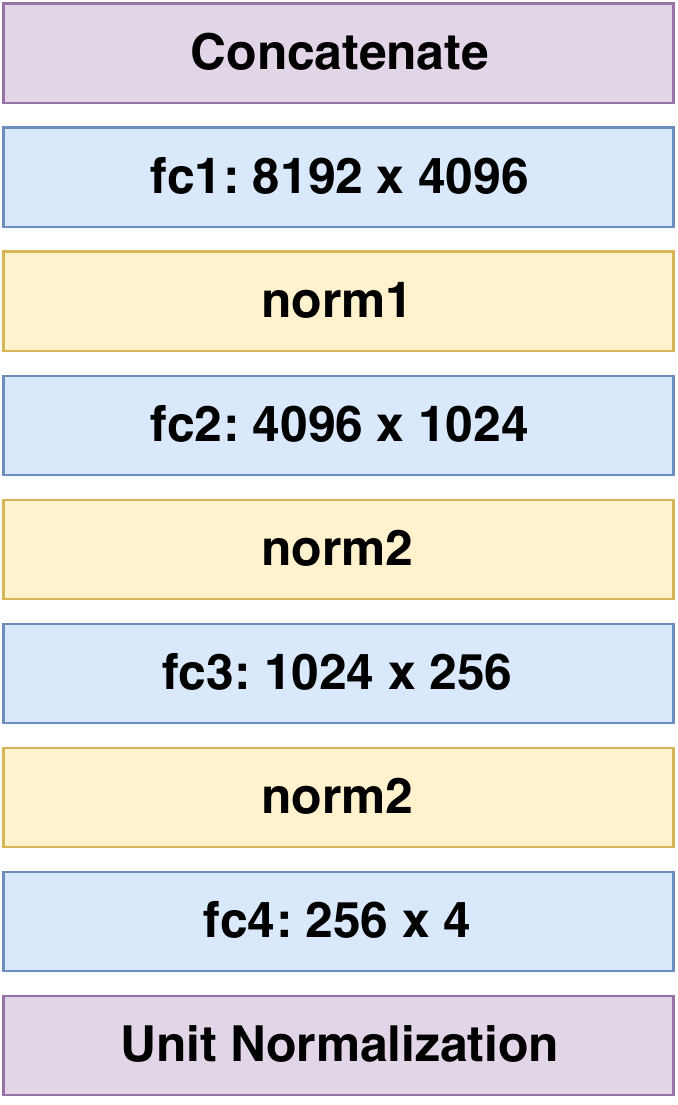}
   \end{center}
   \caption{Architecture of the pose output stage}
   \label{fig:pose_output_stage}
\end{figure}

The first step in the pose refinement stage is to concatenate the outputs of the other two stages. The resultant 8,192 dimensional vector is fed through 4 fully connected layers to get a 4 dimensional quaternion output. We looked into enforcing a constraint on this stage to output unit-normalized quaternions as explored in \cite{leeenforcing}. However, we found that unit-normalizing the 4 dimensional output is simpler and works just as well. \Cref{fig:pose_output_stage} shows the architecture of this stage.

\subsection{Loss Function}
\label{loss}

For our loss function, we attempt to minimize the angle between our output rotation and our sample label. We convert the quaternion to an axis-angle representation of the rotation. The axis is arbitrary. However, the angle is the metric of the error in our pose. Thus, for output $q_{out}$ and label $q_{label}$, we use \cref{eqn:geodesic}:
\begin{equation}
   \label{eqn:loss}
   l = 2\arccos\left(\left|q_{out} * q_{label}^{-1}\right|\right)   
\end{equation}
This loss function disregards the axis of the rotation and only minimizes the angle of the rotation. 

\subsection{Training}
\label{training}

A single training sample consisted of a single bounding box image of an object along with a course estimate of the pose of the object. For training, the estimated pose was generated by rotating the ground truth pose around a random axis by an angle drawn from a uniform distribution:
\begin{multline}
   \label{eqn:axisangle_noise}
   \hat{n}_{noise} = \left(\sqrt{1 - s^2}\cos{\alpha},\sqrt{1 - s^2}\sin{\alpha}, s\right),\\
   \alpha \in \mathcal{U}\left(0, 2\pi\right), s \in \mathcal{U}\left(-1, 1\right);\\
   \theta_{noise} \in \mathcal{U}\left(0, 30\right)   
\end{multline}
Thus
\begin{equation}
   q_{in} = q_{GT} \times q_{noise}
\end{equation}
for $q_{noise}$ generated by using \cref{eqn:axisangle_noise} in \cref{eqn:axisangle_to_quat}. Similarly, the label for that sample is generated by using $\left(\hat{n}_{noise}, -\theta_{noise}\right)$ in \cref{eqn:axisangle_to_quat}. In this manner, a single bounding box image can be used for multiple samples by generating several $q_{in}$ poses. This dataset augmentation is extremely useful when dealing with smaller training datasets. However, we had ample number of samples from the YCB-Video dataset so no such augmentation was required. We did, however, verify that our network did not over-fit to the images by resampling test images with different $q_{noise}$ pose estimates in \cref{network_invarience}.

We first trained our network on 3,000 images for each object. This served as the initialization of our network weights. For testing for each individual object, we fine tuned our network using the remaining 2,000 training images.

Our non-linear loss function (\cref{eqn:loss}) has a number of local minima. This can be seen intuitively since we are outputting a full rotation as a unit quaternion but only minimizing the angle of the rotation. To avoid those local minima while still taking advantage of our loss function, we first trained our network using the MSE loss for 5 epochs, and then continued training with our loss function for another 10 epochs. \cite{mahendran20173d} and \cite{tron2009distributed} both recommended this strategy when using the error angle for loss. We used the Adam optimizer for all training.

\section{Results and Discussions}

In this section, we introduce our evaluation metric, and discuss the experiments that we performed to verify our claims. We also discuss the decisions we made in our framework and verify our choices. We evaluate our network's performance with \begin{enumerate*}[label=(\roman*)]
   \item Varying size of training dataset
   \item Different CNN networks for feature extraction stage
   \item Different input pose estimation distributions
   \item Over-fitting to images
\end{enumerate*}

\subsection{Evaluation Metrics}

There are three standard metrics used to evaluate the performance of 6D pose estimation frameworks -- 2D re-projection error, IoU score, and average 3D distance of model vertices. Each of these metrics provides a threshold for correct estimation of the pose, the final score being the percentage of samples estimated above the threshold. In the 2D re-projection error metric as used by \cite{brachmann2016uncertainty}, the mean distance between the projection of the object's 3D mesh vertices onto the image using the estimate and the ground truth are measured as the error, with a value of 5 pixels taken as the threshold for correctness. Since this metric heavily factors in the object detection and 3D localization task, it was not ideal for us. The IoU score, used by \cite{kehl2017ssd}, is the measure of the overlap between the projected 3D model using the ground truth pose and the predicted pose, with an overlap of 0.5 taken as the threshold. This is, again, not a very useful metric for 3D rotation. The average 3D distance of model vertices, also known as the ADD method, uses the mean distance between the object's 3D mesh vertices at ground truth pose and estimated pose, with the threshold scaling by size. This method, used by \cite{rad2017bb8} and \cite{xiang2018posecnn} among others, is better for 3D pose estimation. However, since our framework does not rely on the 3D model of the object for pose refinement, we decided not to use it.

\cite{mahendran20173d}, who also rely solely on bounding box RGB images for 3D pose estimation, use the geodesic angle between the ground truth and their estimate to evaluate their system. This is simply the angle of an axis-angle rotation required to rotate the estimated pose back to ground truth. Thus, we generated a refined pose from our output unit quaternion as:
\begin{equation}
   q_{refined} = q_{in} * q_{out}
\end{equation}
Then, we calculated the geodesic angle between the refined pose and the ground truth pose using \cref{eqn:geodesic}:
\begin{equation}
   \theta_{error} = 2\arccos{\left(\left|q_{refined} * q_{GT}^{-1} \right|\right)}
\end{equation}
This is the same as our loss function (\cref{eqn:loss}), but between the ground truth pose and the final refined pose rather than between the sample label and the output quaternion. We used this metric for all our evaluations in the section below.

\subsection{Experiments}
We ran several different experiments to verify each decision we made in our framework and to highlight its advantages. All our code was written in Python 3.6 using the PyTorch library. The network was trained on a system with an AMD Ryzen 7 2700X CPU and Nvidia GTX 1080 Ti GPU. The same system was used for testing. Some inference speed testing was done on the CPU instead of the GPU.

\begin{figure}
   \begin{center}
      \includegraphics[width=\linewidth]{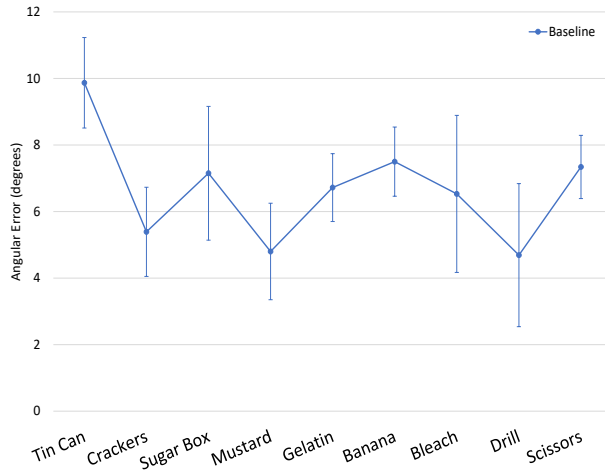}
   \end{center}
   \caption{Baseline performance of our pose refinement framework}
   \label{fig:baseline}
\end{figure}

For each experiment, we first established a baseline performance using the training parameters stated in \cref{training}. The baseline results are shown in \cref{table:baseline} and \cref{fig:baseline}. All results are tabulated as the mean angular error between the refined pose and the ground truth pose. The plots also show the standard deviation.

\begin{table}
   \begin{center}
      \begin{tabular}{|l|p{0.15\linewidth}|p{0.15\linewidth}|}
         \hline
         Baseline Results & Mean Angular Error & Standard Deviation \\
         \hline\hline
         Tin Can & 9.87 & 1.36  \\
         Crackers & 5.39 & 1.34  \\
         Sugar Box & 7.15 & 2.01  \\
         Mustard & 4.80 & 1.45  \\
         Gelatin & 6.72 & 1.02  \\
         Banana & 7.50 & 0.96 \\
         Bleach & 6.53 & 2.36  \\
         Drill & 4.69 & 1.38  \\
         Scissors & 7.34 & 0.95  \\
         \hline
         Average & 6.67 & 1.42  \\
         \hline
      \end{tabular}
   \end{center}
   \caption{Baseline performance of our pose refinement framework}
   \label{table:baseline}
\end{table}

\subsubsection{Varying Training Dataset Size}
The primary advantage of a relatively shallow neural network architecture is that it requires less training with fewer training samples to achieve optimal performance. To verify this, we retrained our network using \begin{enumerate*}[label=(\roman*)]
   \item 5,000 base training samples from each object and 5,000 fine tuning samples for the specific object, and
   \item 1,000 base training samples from each object and 1,000 fine tuning samples for the specific object
\end{enumerate*}, and compared the results against the baseline of 3,000 base base training samples of each class and 2,000 fine tuning samples. The results can be seen in \cref{table:training_size}
\begin{table}
   \begin{center}
      \begin{tabular}{|l|p{0.15\linewidth}|p{0.15\linewidth}|p{0.15\linewidth}|}
         \hline
         Training Samples & Baseline & 5,000 & 1,000\\
         \hline\hline
         Tin Can & 9.87 & 9.21 & 10.64 \\
         Crackers & 5.39 & 4.75 & 6.03  \\
         Sugar Box & 7.15 & 7.09 & 7.96  \\
         Mustard & 4.80 & 4.36 & 5.24  \\
         Gelatin & 6.72 & 6.68 & 8.21  \\
         Banana & 7.50 & 6.35 & 7.98 \\
         Bleach & 6.53 & 6.29 & 7.06  \\
         Drill & 4.69 & 4.62 & 5.68 \\
         Scissors & 7.34 & 7.16 & 8.73  \\
         \hline
         Average & 6.67 & 6.28 & 7.50 \\
         \hline
      \end{tabular}
   \end{center}
   \caption{Perform of our framework with different training dataset sizes}
   \label{table:training_size}
\end{table}

\begin{figure}
   \begin{center}
      \includegraphics[width=\linewidth]{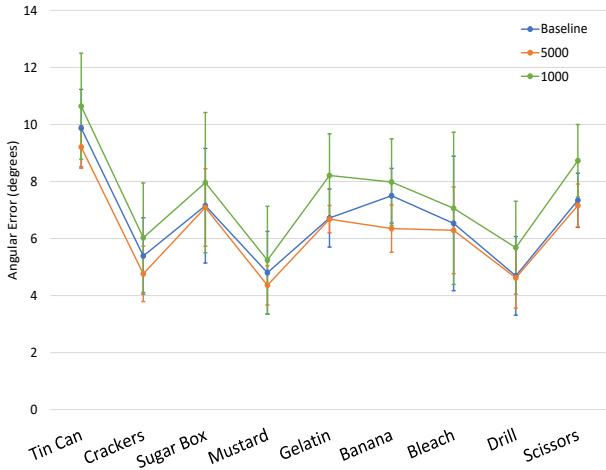}
   \end{center}
   \caption{Perform of our framework with different training dataset sizes}
   \label{fig:training_size}
\end{figure}

Our network retained its performance even with just 9,000 base training samples (1,000 for each object), and just 1,000 fine-tuning samples. Furthermore, increasing the training samples beyond our baseline did not increase the performance significantly. This highlights the clear advantage of a shallow neural network in training.

\subsubsection{Different CNN Feature Extraction Stages}
\label{exp_CNN}

Most neural network architecture literature recommends using CNN networks pretrained on large image datasets such as ImageNet and fine tuned for our particular application. While effective, most such CNN networks are often extremely dense and require long inference times, especially on CPUs. Thus, another advantage of our shallow CNN feature extraction stage is that it can extract the necessary image features for pose refinement without lengthy inference times on constrained hardware. We compared the baseline performance of our framework with our shallow CNN feature extraction stage against a pretrained CNN stage. For this task, we chose the VGG-M network \cite{chatfield2014return} with weights initialized on ImageNet. This network is one of the shallower pretrained VGG networks. We evaluated results both by performing base training on the network and then fine tuning it for each specific object, as well as  simply fine tuning without any base training on our dataset. The other two stages were not changed / retrained in this experiment. We also tested inference time by testing on the CPU.

\begin{figure}
   \begin{center}
      \includegraphics[width=\linewidth]{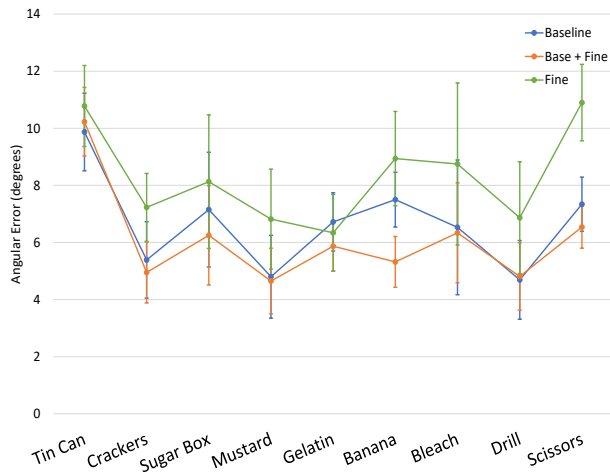}
   \end{center}
   \caption{Performance of our CNN network against VGG-M network}
   \label{fig:cnn_network}
\end{figure}

As can be seen in \cref{table:cnn_network} and \cref{fig:cnn_network}, the VGG-M network performed marginally better than our CNN network in the feature extraction stage with base training. With just fine tuning, our network beat the VGG-M network for most of the objects with a fair margin. More interestingly, our network had significantly faster inference on the CPU. On average, our CNN network in our framework was able to refine poses over thrice as fast as the VGG-M network. We believe that our network provides a valid tradeoff between performance and inference time.

\begin{table}
   \begin{center}
      \begin{tabular}{|l|p{0.15\linewidth}|p{0.15\linewidth}|p{0.15\linewidth}|}
         \hline
         Training Samples & Baseline & VGG-M (Base + Fine) & VGG-M (Fine only) \\
         \hline\hline
         Tin Can & 9.87 & 10.23 & 10.78 \\
         Crackers & 5.39 & 4.95 & 7.23 \\
         Sugar Box & 7.15 & 6.25 & 8.13  \\
         Mustard & 4.80 & 4.65 & 6.82  \\
         Gelatin & 6.72 & 5.87 & 6.34  \\
         Banana & 7.50 & 5.32 & 8.94 \\
         Bleach & 6.53 & 6.34 & 8.75  \\
         Drill & 4.69 & 4.81 & 6.87 \\
         Scissors & 7.34 & 6.54 & 10.90  \\
         \hline
         Average & 6.67 & 6.11 & 8.30 \\
         \hline
      \end{tabular}
   \end{center}
   \caption{Performance of our CNN network against VGG-M network}
   \label{table:cnn_network}
\end{table}

\begin{table}
   \begin{center}
      \begin{tabular}{|l|p{0.15\linewidth}|p{0.15\linewidth}|p{0.15\linewidth}|}
         \hline
         Network & Average FPS & Average Performance \\
         \hline\hline
         Ours & 78 fps & 6.67 \\
         VGG-M (Base + Fine) & 24 fps & 6.67 \\
         VGG-M (Base + Fine) & 24 fps & 8.30 \\
         \hline
      \end{tabular}
   \end{center}
   \caption{Average speed and performance of our CNN network against VGG-M network}
   \label{table:cnn_network}
\end{table}

\subsubsection{Different Input Pose Estimate Distributions}
\label{noise_distribution}
\begin{figure}
   \begin{center}
      \includegraphics[width=\linewidth]{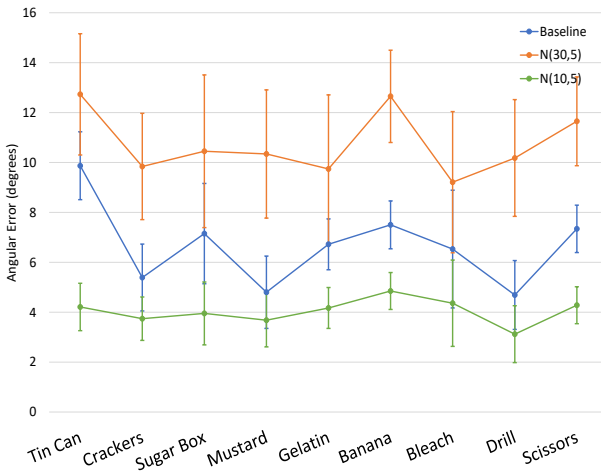}
   \end{center}
   \caption{Performance of our framework with different input pose estimate distributions}
   \label{fig:noise_distribution}
\end{figure}

Our pose refinement framework is flexible enough to be used with any pose estimation framework with varying performance. To verify that it works with an extremely coarse pose estimation system just as well as with a highly precise pose estimation system, we simulated tests with varying pose estimate inputs. For our baseline, pose estimates for our training and testing data was generated using axis-angle rotations with angles drawn from a uniform distribution from $0\degree$ to $30\degree$ (\cref{eqn:axisangle_noise}). We compared this performance against pose estimates with angles drawn from two different normal distributions. The results can be seen in \cref{table:noise_distribution} and \cref{fig:noise_distribution}.

Our framework proved to be extremely robust when given input pose estimates of varying precision. Even a highly precise estimate from an $\mathcal{N}\left(10, 5\right)$ distribution was further refined by our framework, while refinement performance was consistent with a course input pose estimate from an $\mathcal{N}\left(30, 5\right)$ distribution. In both cases, the standard deviation for each pose went down significantly.
\begin{table}
   \begin{center}
      \begin{tabular}{|l|p{0.15\linewidth}|p{0.15\linewidth}|p{0.15\linewidth}|}
         \hline
         Training Samples & Baseline & $\mathcal{N}\left(30,5\right)$ & $\mathcal{N}\left(10,5\right)$\\
         \hline\hline
         Tin Can & 9.87 & 12.73 & 4.21 \\
         Crackers & 5.39 & 9.84 & 3.74 \\
         Sugar Box & 7.15 & 10.45 & 3.95  \\
         Mustard & 4.80 & 10.34 & 3.68  \\
         Gelatin & 6.72 & 9.74 & 4.17  \\
         Banana & 7.50 & 12.65 & 4.85 \\
         Bleach & 6.53 & 9.21 & 4.36  \\
         Drill & 4.69 & 10.18 & 3.12 \\
         Scissors & 7.34 & 11.65 & 4.28  \\
         \hline
         Average & 6.67 & 10.75 & 4.04 \\
         \hline
      \end{tabular}
   \end{center}
   \caption{Performance of our framework with different input pose estimate distributions}
   \label{table:noise_distribution}
\end{table}

\subsubsection{Over-fitting to Images}
\label{network_invarience}

Our network should factor both the bounding box image as well as the input pose estimate into the final refined pose. A challenge for our network architecture is to ensure that the network does not over-fit to training images. Our solution to this challenge was to have our network output a refining rotation rather than a final refined pose. To verify this design choice, we tested our network by resampling each of our testing images multiple times with a different pose estimate for each sample. Each pose estimate was created by generated using axis-angle rotations with angles drawn from a uniform distribution from $0\degree$ to $30\degree$ (\cref{eqn:axisangle_noise}). 

\begin{table}
   \begin{center}
      \begin{tabular}{|l|p{0.15\linewidth}|p{0.15\linewidth}|}
         \hline
         Training Samples & Baseline & With Resampling\\
         \hline\hline
         Tin Can & 9.87 & 9.57 \\
         Crackers & 5.39 & 5.24  \\
         Sugar Box & 7.15 & 7.34  \\
         Mustard & 4.80 & 4.96  \\
         Gelatin & 6.72 & 6.78  \\
         Banana & 7.50 & 7.38 \\
         Bleach & 6.53 & 6.87  \\
         Drill & 4.69 & 4.96  \\
         Scissors & 7.34 & 7.24  \\
         \hline
         Average & 6.67 & 6.70  \\
         \hline
      \end{tabular}
   \end{center}
   \caption{Performance of our framework with resampling of images}
   \label{table:network_invarience}
\end{table}

\begin{figure}
   \begin{center}
      \includegraphics[width=\linewidth]{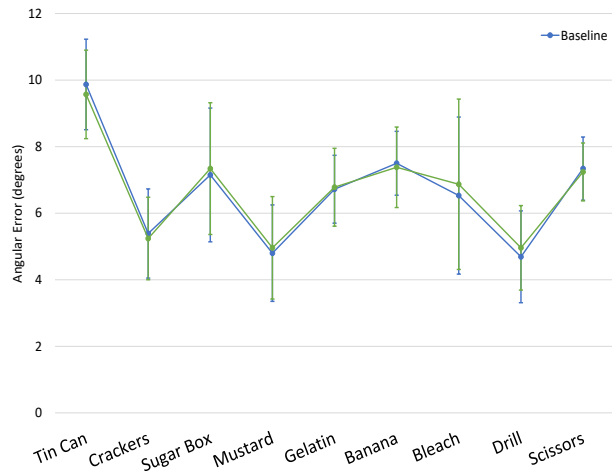}
   \end{center}
   \caption{Performance of our framework with resampling of images}
   \label{fig:network_invarience}
\end{figure}

As can be seen in \cref{table:network_invarience} and \cref{fig:network_invarience}, the performance of our network does not drop with resampled images. This shows that our network does not over-fit to images, and factors both the image as well as the estimated pose equally in its final output.
\section{Conclusion}

We have proposed a new 3D pose refinement framework that can be used in conjunction with a course pose estimation framework to produce highly precise object pose estimates with minimal training data. Through our experiments, we have shown that our framework is robust, fast and precise. We have also used these experiments to justify several of our design choices.
\clearpage

{\small
\bibliographystyle{ieee}
\bibliography{egbib}
}

\end{document}